\documentclass{article}

\usepackage[numbers]{natbib}

\usepackage[preprint]{nips_2018}
\usepackage[utf8]{inputenc} 
\usepackage[T1]{fontenc}    
\usepackage{hyperref}       
\usepackage{url}            
\usepackage{booktabs}       
\usepackage{amsfonts}       
\usepackage{nicefrac}       
\usepackage{microtype}      
\usepackage{graphicx}
\usepackage{subcaption}
\usepackage{mwe}
\usepackage{placeins}

\title{Reproducibility Report for <Learning To Count Objects In Natural Images
For Visual Question Answering>}

\author{
  Shagun Sodhani\thanks{Equal Contribution} \\
    MILA\\
    D\'epartement d'informatique et de recherche op\'erationnelle\\
    Universit\'e de Montr\'eal \\
  \texttt{sshagunsodhani@gmail.com} \\
  \And
    Vardaan Pahuja\footnotemark[1] \\
    MILA\\
    D\'epartement d'informatique et de recherche op\'erationnelle\\
    Universit\'e de Montr\'eal \\
  \texttt{vardaanpahuja@gmail.com} \\
}

\begin{document}

\maketitle

\section{Paper Summary}

Most of the visual question-answering (VQA) models perform poorly on the task of counting objects in an image. The reasons are manifold:

\begin{enumerate}
    \item Most VQA models use a soft attention mechanism to perform a weighted sum over the spatial features to obtain a single feature vector. These aggregated features help in most categories of questions but seem to hurt for counting based questions.
    \item For the counting questions, we do not have a ground truth segmentation of where the objects to be counted are present on the image. This limits the scope for supervision.
\end{enumerate}

Additionally, we need to ensure that any modification in the architecture, to enhance the performance on the counting questions, should not degrade the performance on other classes of questions.

The paper proposes to overcome these challenges by using the attention maps (and not the aggregated feature vectors) as the input to a separate \textbf{count} module. The basic idea is quite intuitive: when we perform weighted averaging based on different attention maps, we end up averaging the features corresponding to the different instances of an object. This makes the feature vectors indistinguishable from the scenario where we had just one instance of the object in the image. Even multiple glimpses (multiple steps of attention) can not resolve this problem as the weights given to one feature vector would not depend on the other feature vectors (that are attended to). Hard attention could be more useful than soft-attention but there is not much empirical evidence in support of this hypothesis. 

\subsection{Motivation}
The motivation for the proposed approach comes directly from one of the limitations of the existing approaches which is to use features based on soft-attention for all classes of questions, including the counting questions. The paper proposed to use a separate class of counting features which are obtained from the raw attention maps (and not the aggregated feature maps). To ensure that their model is plug-and-play, the authors designed the count features such that the module using them can be integrated into any VQA model (that uses attention based features). The authors also aimed to ensure that any new module, that is added to the original architecture, should not degrade the performance of the original architecture. With these motivations, the authors formed separate counting features by using the soft-attention maps and the bounding box proposals and integrated these features into the main classifier. In the ideal setting, the output of the counting component would be a one-hot vector with a non-zero value at the index that corresponds to the count of the object in the image. This clearly aids the VQA model in predicting the correct answer for count-type questions.

\subsection{Proposed approach}

The proposed \textbf{count} module is a separate pipeline that can be integrated with most of the existing attention based VQA models, without affecting the performance on non-count based questions. The inputs to the \textbf{count} module are the attention maps and the object proposals (obtained from some pre-trained model like the R-CNN model) and the output is a count-feature vector which is used to answer the count-based question. The top level idea is the following - given the object proposals and the attention maps, create a graph where nodes are objects (object proposals) and edges capture how similar two object proposals are (how much do they overlap). The graph is transformed (by removing and scaling edges) so that the true count of the number of underlying objects can be obtained easily.

The inputs to the \textbf{count} module are the object proposals (in terms of bounding boxes) and attention weights for these proposals based on the question. The attention weights are such that, more is the relevance of a proposal, more is its attention weight. To explain their methodology, the paper simplifies the setting by making two assumptions. The first assumption is that the attention weights are either 1 (when the object is present in the proposal) or 0 (when the object is absent from the proposal). The second assumption is that any two object proposals either overlap completely (in which case, they are corresponding to the exact same object and hence receive the exact same weights) or the two proposals have zero overlap (in which case, they must be corresponding to completely different objects). These simplifying assumptions are made only for the sake of exposition and do not limit the capabilities of the \textbf{count} module.

Given the assumptions, the task of the \textbf{count} module is to handle the exact duplicates to prevent double-counting of objects. As the first step, the attention weights ($a$) are used to generate an attention matrix ($A$) by performing an outer product between $a$ and $a^{T}$. This corresponds to the step of creation of a graph from the input. $A$ corresponds to the adjacency matrix of that graph. The attention weight for the $i^{th}$ proposal corresponds to the $i^{th}$ node in the graph, and the edge between the nodes $i$ and $j$ has the weight $a_i*a_j$. Also note that the graph is a weighted directed graph and the subgraph of vertices satisfying the condition $a_i = 1$ is a complete directed graph with self-loops. Given such a graph, the number of vertices is $|V|$ = $\sqrt{|E|}$. $|E|$ could be computed by summing over the adjacency matrix. This implies that if the proposals are distinct, the count can be obtained trivially by performing a sum over the adjacency matrix. 

The objective is now to eliminate the edges such that the underlying objects are the vertices of a complete subgraph. This requires removing two type of duplicate edges - intra-object edges and inter-object edges. 

Intra-object edges can be removed by computing a distance matrix, $D \in \mathcal{R}^{n \times n}$, defined as $D_{ij} = (1 - \textrm{IoU}(b_i,b_j))$, where $\textrm{IoU}$ matrix corresponds to the Intersection-over-Union matrix and $b_i$ corresponds to the bounding-box of the $i^{th}$ proposal. A modified adjacency matrix $\tilde{A}$ is obtained by performing the element-wise product between $f_1(A)$ and $f_2(D)$, where $f_1$ and $f_2$ are piece-wise linear functions that are learnt via backpropogation.

The inter-object edges are removed in the following manner: Count the number of proposals that correspond of each instance of an object and then scale down the edges corresponding to the different instances by that number. This creates the effect of reducing the weights of multiple proposals equivalent to a single proposal. The number of proposals corresponding to an object is not available as an annotation in the training pipeline and is estimated based on the similarity between the different proposals (measured via the attention weights $a$, adjacency matrix $A$ and distance matrix $D$). The matrix corresponding to the similarity between proposals  ($sim_{i, j}$) is transformed into a vector corresponding to the scaling factor of each node ($s_i$) as shown below:

\begin{equation}
    s_i = \frac{1}{\sum_{j}sim_{i, j}}
\end{equation}

$s$ can be converted into a matrix (by doing outer-product with itself) so as to scale both the incoming and the outgoing edges. The self edges (which were removed while computing $\tilde{A}$) are added back (after scaling with $s$) to obtain a new transformed matrix $C$ as given below:

\begin{equation}
    C = \tilde{A} \odot s s^T + \textrm{diag}(s \odot f_1(a \odot a))
\end{equation}

The transformed matrix $C$ is a complete graph with self-loops where the nodes corresponds to all the relevant object instances and not the object proposals. The actual count can be obtained from $C$ by performing a sum over all its values as described earlier. The original count problem was a regression problem but it is transformed into a classification problem to avoid scale issues. The network produces a $k$-hot $n$-dimensional vector $o$, where $n$ is the number of object proposals that were fed into the module (and hence the upper limit on how large a number could the module count). In the ideal setting, $k$ should be one, as the network would produce an integer value but in practice, the network produces a real number so $k$ can be upto 2. If $c$ is an exact integer, the output is a 1-hot vector with the value in index corresponding to $c$ set to 1. If $c$ is a real number, the output is a linear interpolation between two one-hot vectors (the one-hot vectors correspond to the two integers between which $c$ lies).

The \textbf{count} module supports computing the confidence of a prediction by defining two variables $p_a$ and $p_D$ which compute the average distance of $f_6(a)$ and $f_7(D)$ from 0.5. The final output $o^{~}$ is defined as $f_8(p_a + p_D) . o$

All the different $f$ functions are piece wise linear functions and are learnt via backpropagation.

\section{Experiments}

\subsection{Motivation for the experiments}

The authors motivated their approach by making two simplifying assumptions (attention weights are either 1 or 0 and any two object proposals either overlap completely or have zero overlap). In this ideal scenario, the authors argue that their \textbf{count} module would be able to exactly compute the number of instances of the relevant objects given an image. In the real life scenario, these assumptions do not hold exactly and empirical evidence is needed to evaluate the performance of the model on real data. 

The authors have argued that their \textbf{count} module could be used with most of the VQA models. Since they claim novelty on the use of \textbf{count} module (and not on what base VQA model is used), they used the model proposed by \cite{kazemi2017show} as the base model without making any changes to its architecture or the hyper parameters. It seems to be a justified choice given that the base model achieved the state-of-the-art results on VQA v1 dataset. This model is augmented with the \textbf{count} module, which uses object feature proposals as an additional feature. These object proposals come from the Faster R-CNN model \cite{2015arXiv150601497R} (as used by \cite{anderson2017bottom}).

The authors also performed experiments on a toy dataset to understand the intricacies of the \textbf{count} module. Since those experiments are for exposition and not for evaluating the performance of the \textbf{count} module, we did not reproduce those experiments.

\subsection{Experiments demonstrated in the paper}

The core problem that the paper proposes to solve is the following: improving the performance of VQA models on count based questions without degrading the performance on other category of questions. To provide empirical evidence for the effectiveness of their solution, the paper evaluates the VQA model (augmented with the \textbf{count} module on the validation set, test-dev and test set of VQA v2 \cite{2016arXiv161200837G}. The performance of this \textbf{count} augmented VQA model is compared with the performance of the base VQA model on count-based questions (to verify that the use of \textbf{count} module helps with count-based questions) and on other class of questions like `Yes/No', `Other' etc (to verify that the use of \textbf{count} module does not deteriorate the performance on other questions). The answers are evaluated according to the following accuracy metric:

$$ accuracy = min\big(\frac{\mbox{no.of humans who gave that answer}}{3}, 1\big) $$

The original VQA v2 dataset provides a category of questions called as `Number' questions which contains both count-based questions as well as questions like ``what time is it?'' for which the answer is also a number. Since the \textbf{count} module is designed specifically to handle the count-based questions and not the number-based questions, the  paper introduces a new sub-category of `Count' questions in which they consider only questions starting with `How many' and ignore the other number-based questions. This introduces a slight discrepancy in the evaluation on the test data as the official evaluation server (URL: \url{https://evalai.cloudcv.org/web/challenges/challenge-page/80/phases}) does not support this new sub-category of count-based questions.

The authors additionally evaluate the accuracy of their model on the balanced pairs (which comprise of a single question on two different images). The model must answer correctly on both to receive a score. Evaluation on balanced pairs is important it because the model should learn to find the subtle details between images, instead of being able to exploit the question biases in the dataset.

\subsection{Reproducing the main results}
We implemented both the baseline architecture and the count module from scratch using the details mentioned in the paper and the hyper-parameters suggested by the paper to reproduce the experiments on VQA-v2 dataset. We also implemented the baseline architecture, which does not use a separate count module. We report the results on test-dev and test, as obtained from the official evaluation server in Table \ref{tab:vqa-test}. For the validation set, we report results on individual question categories (in Table \ref{tab:vqa-val}) as well as balanced pairs by using the evaluation script provided by the authors for our trained model (in Table \ref{tab:vqa-val}). We also report the results on validation set as obtained from the official evaluation scripts released by \cite{2016arXiv161200837G}. In a nutshell, our experiments achieve similar level of performance gains, compared to the baseline model, as reported in the paper. For a detailed discussion on this aspect, refer to Section~\ref{sec::discussion}. 

\subsection{Assumptions in implementation}
\label{subsection::assumptions}

\begin{itemize}
    \item The authors proposed a slight modification in the baseline architecture. In the baseline model, vision features $x$ and question features $y$ are fused using the concatenation operation, followed by linear projection and ReLU. The authors replace this operation by a different fusion operation  $x \diamond y= \textrm{ReLU}(W_x x + W_y y) - (W_x x - W_y y)^2$. We used this modified formulation, but it was unclear what is the nature of $W_x$ for vision features. Since a feedforward linear layer is not suitable for image feature maps, we used $1\times1$ convolutional layer, which worked well in our case.
    \item  In the classifier, the authors have mentioned that their model uses both BatchNorm and Dropout, which is seldom recommended in literature but works in this case.
    \item In all of our LSTM variants of different models, we use the cell state of the final timestamp as the feature representation of the question, and the baseline paper doesn't clearly mention if it uses the hidden state or cell state.
    \item We use the attention vectors from the first glimpse as input to the count module. We also try one experiment with use of second glimpse instead of first glimpse attention vector.
    \item The paper does not specify if they have used the validation data as part of training data when evaluating the model over test data. We have used only the training data when evaluating the performance over both validation and test data.
\end{itemize}
   
All other hyper-parameters like learning rate, half-life of LR scheduler (in terms of no. of iterations), type of optimizer (Adam) were clearly mentioned in the paper, and we stick to using these for our experiments.

\subsection{Additional Experiments}

In addition to the experiments proposed in the paper, we performed some more experiments to perform a more holistic ablation study on the choice of different hyper-parameters or sub-modules of the model proposed by the authors. The following additional experiments were performed: 

\begin{itemize}
\item Increasing the number of objects to 20 instead of default 10.
\item Use of unidirectional LSTM instead of unidirectional GRU to obtain the question features.
\item Setting the threshold value for confidence of prediction to 0.2 instead of 0.5.
\item Setting the embedding dimension corresponding to the words to 100 instead of 300.
\item Use of second glimpse attention vector instead of first glimpse attention vector.
\item Use of bidirectional LSTM instead of the unidirectional LSTM for baseline model.
\end{itemize}

\setlength\tabcolsep{3.8pt}
\begin{table}[!ht]
    \caption{
        Results on VQA v2 test and test-dev (obtained from standard evaluation server)
    }
    \label{tab:vqa-test}
    \centering
    \begin{tabular}{l c c c c c c c c}
        \\
        \toprule
        & \multicolumn{4}{c}{VQA v2 test-dev} & \multicolumn{4}{c}{VQA v2 test}\\
        \cmidrule(l{4pt}){2-5} \cmidrule(l{4pt}){6-9}
        Model & Yes/No & \textbf{Number} & Other & All & Yes/No & \textbf{Number} & Other & All \\
        \hline
        Baseline + count (author) & 83.14 & 51.62 & 58.97 & 68.09 & 83.56 & 51.39 & 59.11 & 68.41 \\
        \hline
        \textbf{Baseline + count (ours)} & 81.46 & 47.9 & 56.85 & 66.08 & 81.07 & 47.59 & 56.65 & 65.68 \\
        Unidirectional LSTM & 81.2 & 48 & 56.59 & 65.8 & 81.62 & 47.38 & 56.87 & 66.1 \\
        Use 20 objects & 80.67 & 48.48 & 56.54 & 65.56 & 81.14 & 48.53 & 56.73 & 65.96 \\
        Embed. size 100 & 78.33 & 47.97 & 56.14 & 64.35 & 78.71 & 47.94 & 56.43 & 64.74 \\
        Confidence thresh. 0.2 & 80.59 & 48.78 & 56.72 & 65.66 & 80.9 & 48.85 & 56.86 & 65.96 \\
        Use second glimpse & 80.3 & 48.82 & 56.68 & 65.52 & 80.61 & 48.47 & 56.94 & 65.83 \\
        \hline
        Baseline (author) & 82.98 & 46.88 & 58.99 & 67.50 & 83.21 & 46.60 & 59.20 & 67.78 \\
        \textbf{Baseline (ours)} & 80.67 & 44.76 & 56.45 & 65.1 & 80.75 & 43.76 & 56.76 & 65.29 \\
        Baseline (bidirect. LSTM) & 80.68 & 44.35 & 56.63 & 65.14 & 81.09 & 44.01 & 56.73 & 65.44 \\
        \bottomrule
    \end{tabular}
\end{table}

\setlength\tabcolsep{4.2pt}
\begin{table}[!ht]
    \caption{
        Results on the VQA v2 validation set with models trained only on the training set (uses the answer post-processing as described by the author).
    }
    \label{tab:vqa-val}
    \centering
    \begin{tabular}{l c c c c c c}
        \\
        \toprule
        & \multicolumn{3}{c}{VQA accuracy} & \multicolumn{3}{c}{Balanced pair accuracy} \\
        \cmidrule(l{4pt}){2-4} \cmidrule(l{4pt}){5-7}
        Model & Number & \textbf{Count} & All & Number & \textbf{Count} & All \\
        \hline
        Baseline + count (author) & 49.36$\pm$0.1 & 57.03$\pm$0.0 & 65.42$\pm$0.1 & 23.10$\pm$0.2 & 26.63$\pm$0.2 & 37.19$\pm$0.1 \\
        \hline
        \textbf{Baseline + count (ours)} & 47.31 & 54.65 & 64.25 & 20.96 & 24.18 & 35.66\\
        Unidirectional LSTM & 47.36 & 54.67 & 64.35 & 21.05 & 24.26 & 35.75 \\ 
        Use 20 objects & 48.19 & 55.73 & 64.32 & 21.67 & 25.03 & 35.6 \\
        Embed. size 100 & 47.10 & 54.41 & 63.19 & 20.72 & 23.90 & 33.82 \\
        Confidence thresh. 0.2 & 48.18 & 55.75 & 64.27 & 22.03 & 25.36 & 35.59 \\
        Use second glimpse & 48.04 & 55.57 & 64.2 & 21.83 & 25.27 & 35.49 \\
        \hline
        Baseline (author) & 44.83$\pm$0.2 & 51.69$\pm$0.2 & 64.80$\pm$0.0 & 17.34$\pm$0.2 & 20.02$\pm$0.2 & 36.44$\pm$0.1 \\
        \textbf{Baseline (ours)} & 43.56 & 50.13 & 63.54 & 15.98 & 18.48 & 34.54 \\
        Baseline (bidirect. LSTM) & 43.64 & 50.33 & 63.59 & 15.78 & 18.28 & 34.6 \\
        \bottomrule
    \end{tabular}
\end{table}

\setlength\tabcolsep{3.8pt}
\begin{table}[!ht]
    \caption{
        Results on VQA v2 validation set (obtained from standard evaluation server)
    }
    \label{tab:vqa-val-standard}
    \centering
    \begin{tabular}{l c c c c}
        \\
        \toprule
        Model & Yes/No & \textbf{Number} & Other & All \\ \hline
        \textbf{Baseline + count (ours)} & 80.76 & 47.3 & 55.99 & 64.16 \\
        Unidirectional LSTM & 80.85 & 47.34 & 56.09 & 64.25 \\
        Use 20 objects & 80.46 & 48.19 & 56.11 & 64.22 \\
        Embed. size 100 & 78.54 & 47.06 & 55.59 & 63.09 \\
        Confidence thresh. 0.2 & 80.43 & 48.15 & 56.06 & 64.18 \\
        Use second glimpse & 80.23 & 48.02 & 56.08 & 64.1 \\
        \hline
        \textbf{Baseline (ours)} & 80.23 & 43.52 & 55.92 & 63.43 \\
        Baseline (bidirect. LSTM) & 43.62 & 80.31 & 55.96 & 63.49 \\
        \bottomrule
    \end{tabular}
\end{table}

\section{Discussion}
\label{sec::discussion}

The key idea of the paper is to design and evaluate an independent \textbf{count} module which can be easily incorporated into the existing VQA models and could improve the accuracy of count-based questions, without affecting the other categories. So, we focus our analysis only on the `Count' and `Number' categories. In the paper, all models are run for 100 epochs (which takes over 1 hour per epoch). Due to computational constraints (and given the number of experiments we perform), we run each model only for 30 epochs. We note that the validation accuracy appears to saturate before 30 epochs itself.

\begin{figure}[!ht]
\centering
\includegraphics[scale=0.6]{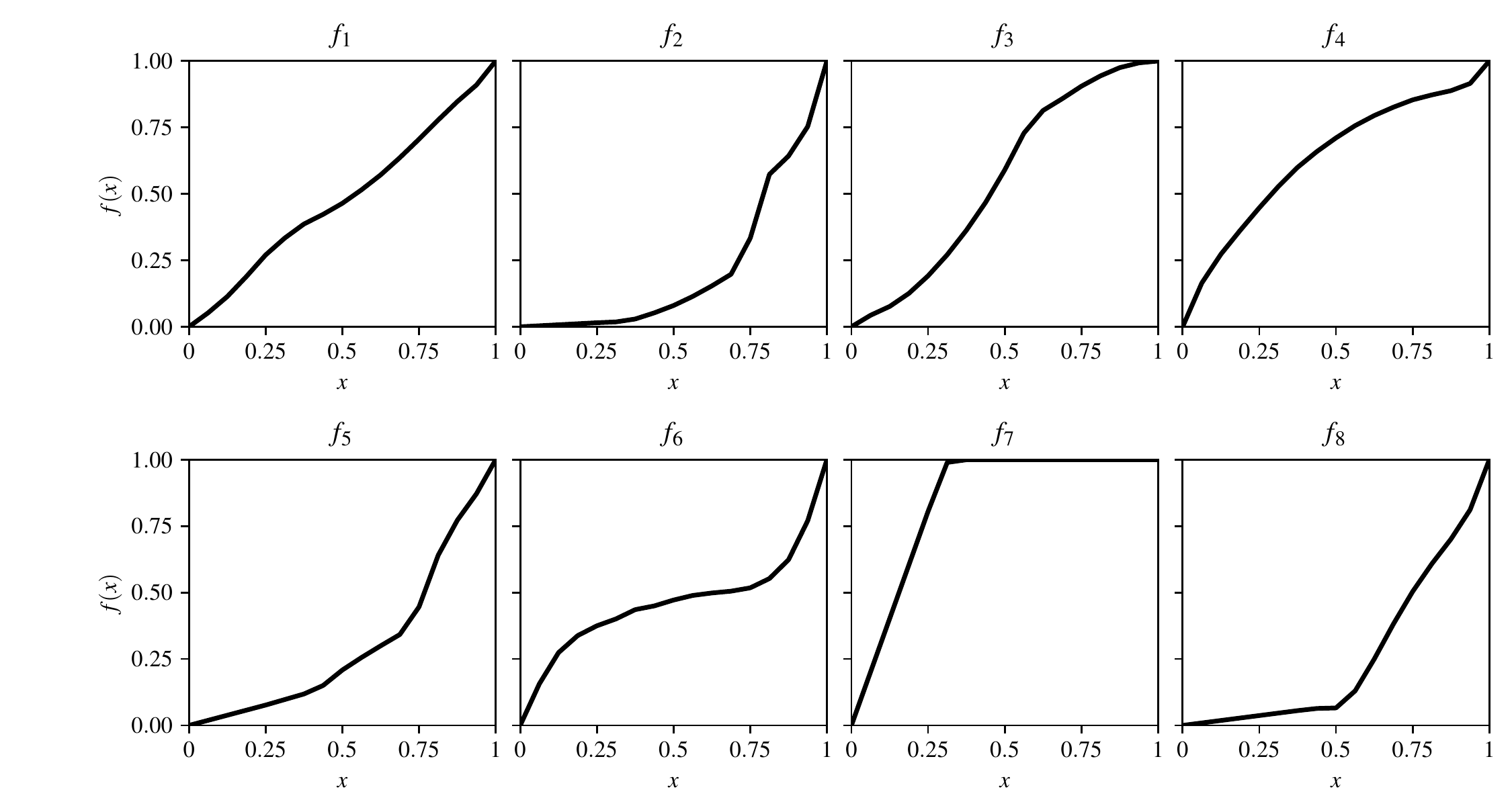}
\caption{Shape of activation functions for (Baseline count) model}
\label{baseline_count_f}
\end{figure}

\begin{figure}[!ht]
\centering
\includegraphics[scale=0.6]{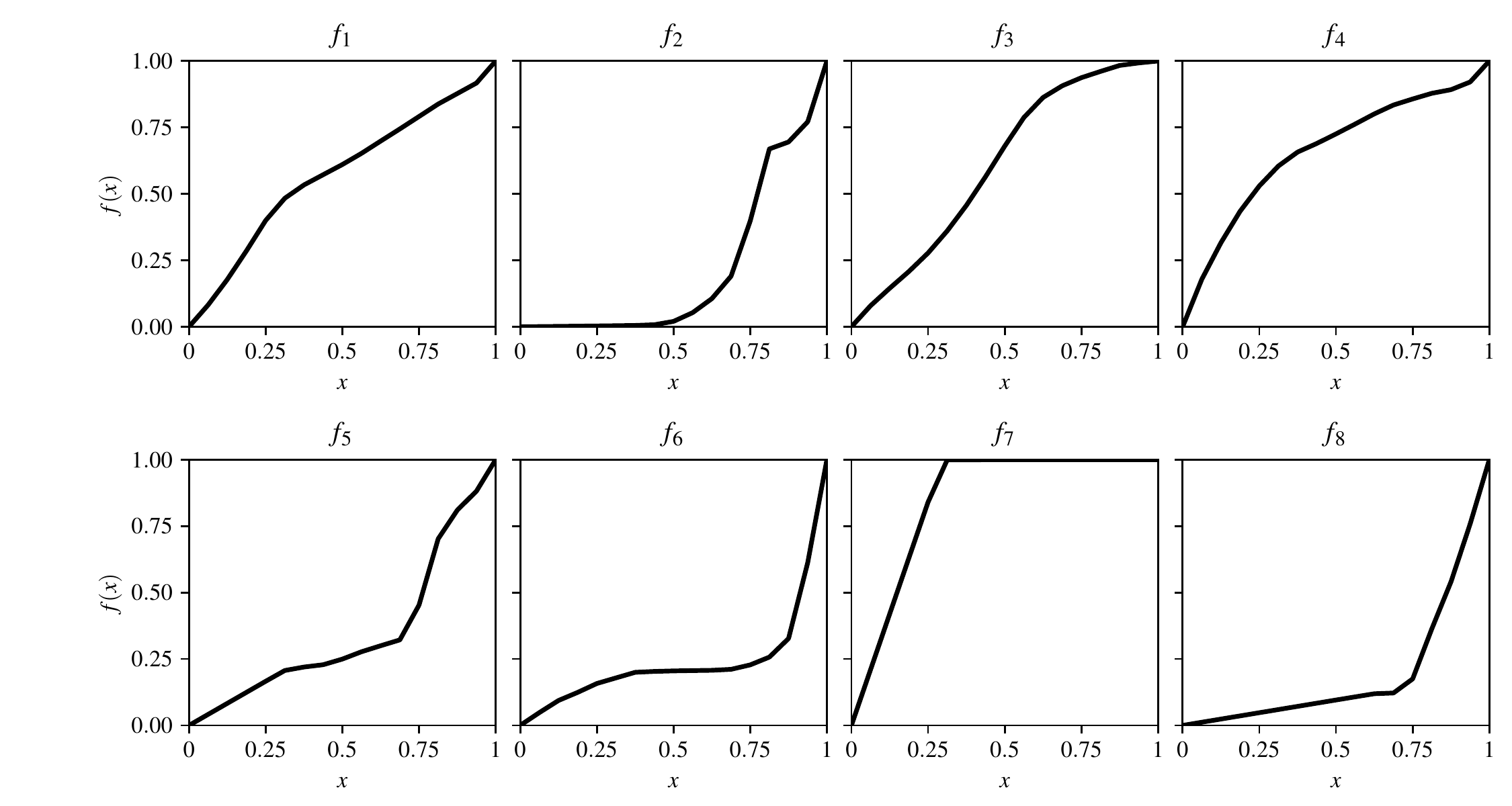}
\caption{Shape of activation functions for (Baseline + count) model with confidence threshold 0.2}
\label{baseline_count_02_f}
\end{figure}

The accuracy for `Number' category obtained using our implementation is 47.9 (test-dev) while the accuracy reported by the paper is 51.62 (test-dev). A possible reason for this mismatch could be that our model is trained for less number of epochs, though we don't expect the accuracy to rise significantly after 30 epochs. Another possibility is that the paper used more than 10 object proposals. The maximum number of object proposals could be up to 100, and the paper reports that a natural choice would be to use 10 object proposals. To validate this assumption, we ran an experiment where the number of object proposals considered in \textbf{count} module, is set to 20, and we do observe a slight improvement in accuracy to 48.48 from the initial value of 47.9. In the model with confidence threshold changed to 0.2 (from 0.5), the accuracy figure for `Number' improves slightly. This is somewhat counter-intuitive as we would expect the model to be completely robust to the value of threshold as $f_6$ and $f_7$ can adjust their parameters as per the threshold. Figure~\ref{baseline_count_f} and Figure~\ref{baseline_count_02_f} compare the two scenarios where the threshold values are set to 0.5 (as proposed by the authors) and 0.2 respectively. While there is no noticeable change in the plot of $f_7$, the plot corresponding to $f_6$ seems to be somewhat shifted. An exact linear shift is not observed as $f_6$ is only a piece-wise linear function. When we reduced the embedding size for question words, we observe degradation in performance on all categories, except the `Number' category. The use of second glimpse as the attention vector input to the count module improves the `Number' accuracy slightly but the overall accuracy is reduced.

For the baseline model \cite{kazemi2017show}, we obtain an accuracy figure of 44.76 which is below the reported figure of 46.88. Use of bidirectional LSTM for the baseline model doesn't make a significant difference. We report the results on validation set from both this paper's evaluation script and the standard evaluation scripts (released with the dataset). The results on the standard evaluation scripts are just slightly lower than the authors' evaluation script, but the difference is not significant.

\section{Conclusion}

The paper proposes a simple and intuitive approach to improve performance over count-based questions for the task of Visual Question Answering. Our motivation behind this report was to judge the work on the parameters of reproducibility. The paper is well written, rich in details  and largely unambiguous. We had to make very few assumptions while implementing the paper (the same have been described in section \ref{subsection::assumptions}. We note that we could not exactly reproduce the results in the paper, but the more likely reason is the limit in terms of computational resources. Our results provide empirical evidence of the effectiveness of the proposed model.

We further performed a series of ablation studies to understand the finer details of the model and bring out the strengths and shortcomings of the paper. Our assessment shows that the model is largely robust to choice of different hyper-parameters and gives significant gains in performance over count-based questions, compared to the baseline model. This can be attributed to the solid analytical foundation of the proposed approach in terms of basic graph theory and the corresponding feature engineering.

\FloatBarrier
\bibliographystyle{plain}
\bibliography{citation}

\begin{thebibliography}{1}

\bibitem{anderson2017bottom}
Peter Anderson, Xiaodong He, Chris Buehler, Damien Teney, Mark Johnson, Stephen
  Gould, and Lei Zhang.
\newblock Bottom-up and top-down attention for image captioning and vqa.
\newblock {\em arXiv preprint arXiv:1707.07998}, 2017.

\bibitem{2016arXiv161200837G}
Y.~{Goyal}, T.~{Khot}, D.~{Summers-Stay}, D.~{Batra}, and D.~{Parikh}.
\newblock {Making the V in VQA Matter: Elevating the Role of Image
  Understanding in Visual Question Answering}.
\newblock {\em ArXiv e-prints}, December 2016.

\bibitem{kazemi2017show}
Vahid Kazemi and Ali Elqursh.
\newblock Show, ask, attend, and answer: A strong baseline for visual question
  answering.
\newblock {\em arXiv preprint arXiv:1704.03162}, 2017.

\bibitem{2015arXiv150601497R}
S.~{Ren}, K.~{He}, R.~{Girshick}, and J.~{Sun}.
\newblock {Faster R-CNN: Towards Real-Time Object Detection with Region
  Proposal Networks}.
\newblock {\em ArXiv e-prints}, June 2015.

\end{thebibliography}

\FloatBarrier
\section{Appendix}

We list some additional figures and plots in the appendix.
\begin{figure}[!ht]
\centering
\includegraphics[scale=0.6]{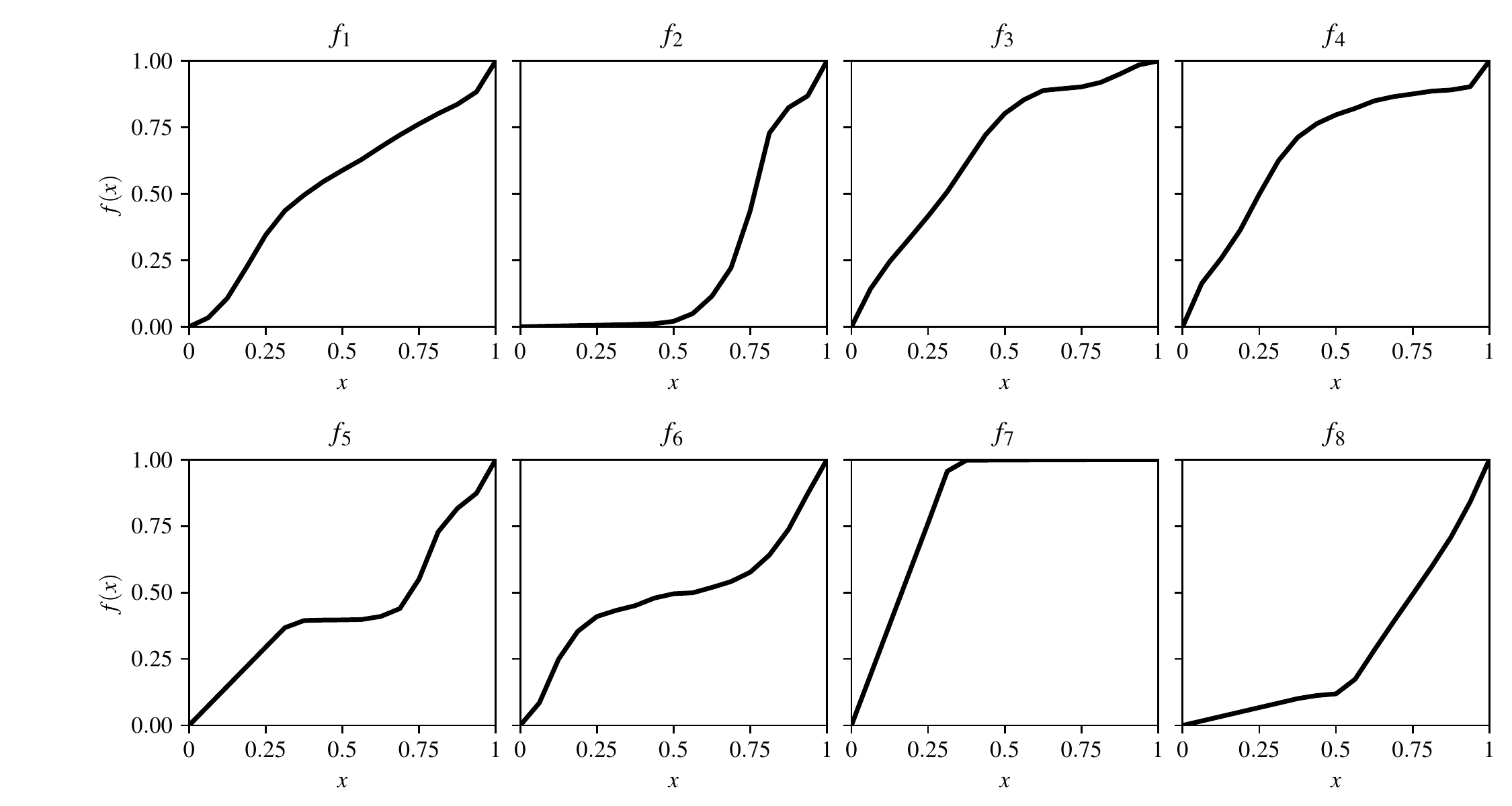}
\caption{Shape of activation functions for (Baseline + count) model with unidirectional LSTM}
\label{baseline_count_lstm_f}
\end{figure}

\begin{figure}[!ht]
\centering
\includegraphics[scale=0.6]{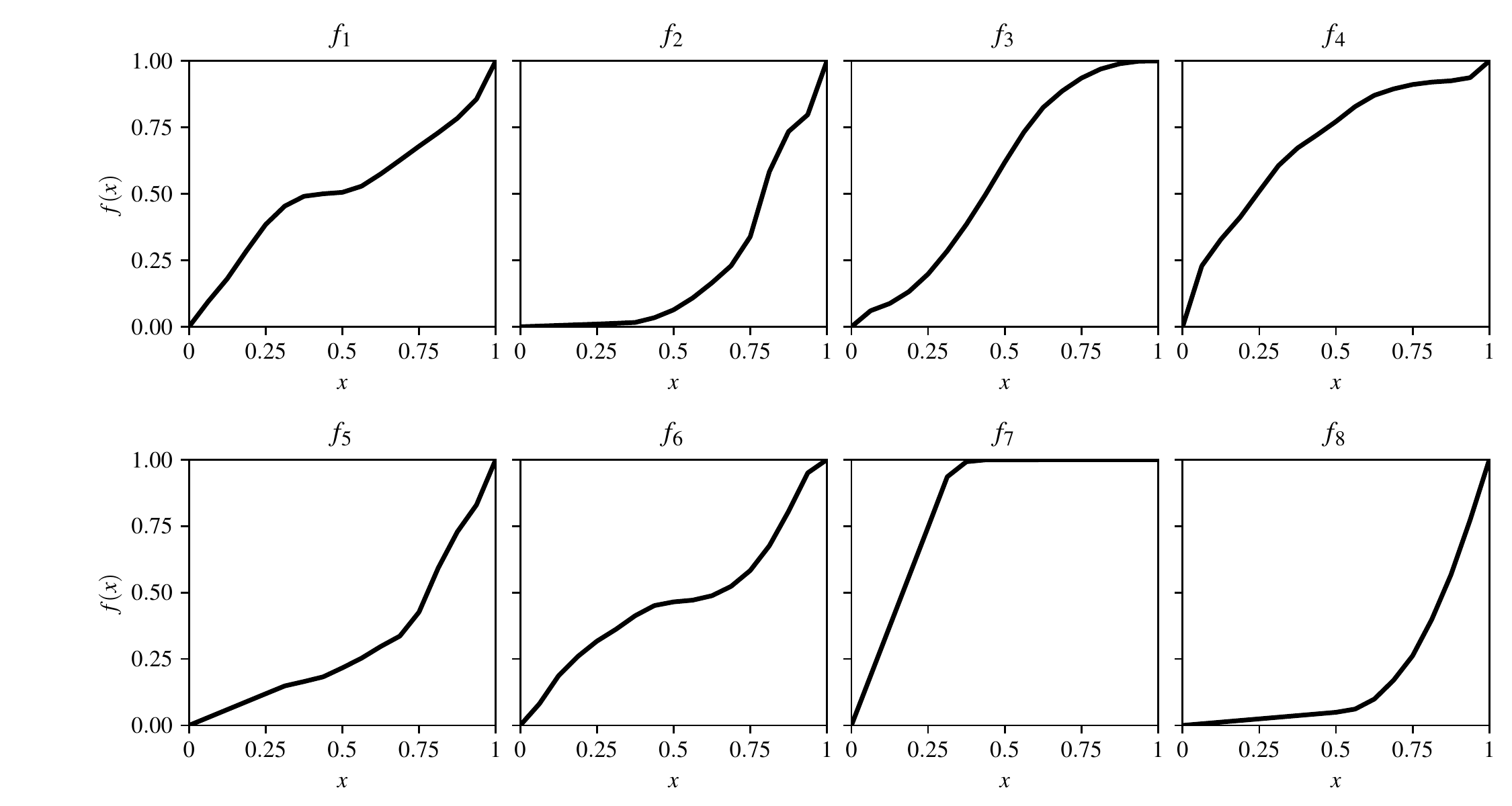}
\caption{Shape of activation functions for (Baseline + count) model with $n=20$ objects}
\label{baseline_count_20_f}
\end{figure}

\begin{figure}[!ht]
\centering
\includegraphics[scale=0.6]{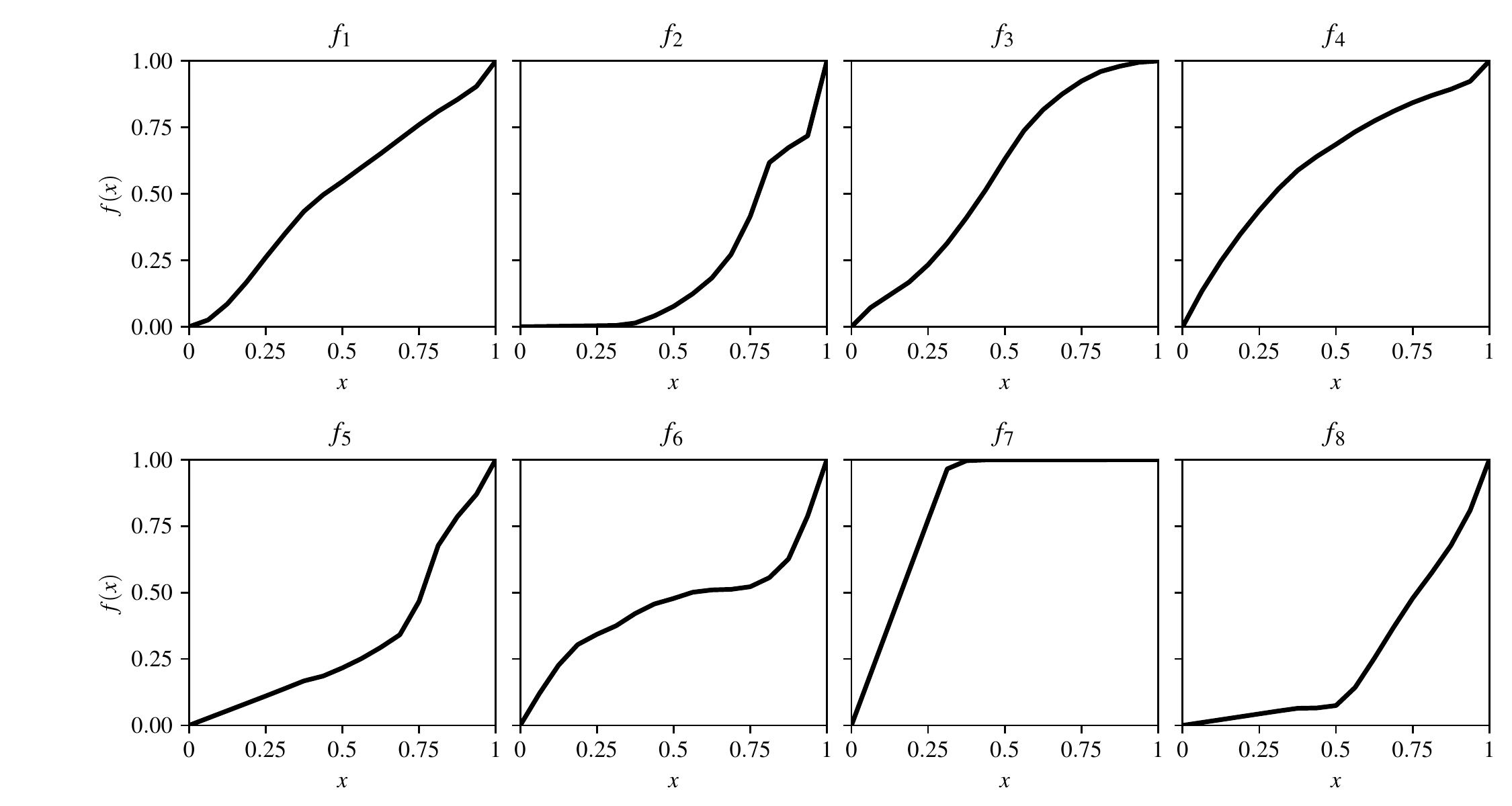}
\caption{Shape of activation functions for (Baseline + count) model with embedding size 100}
\label{baseline_count_embed_f}
\end{figure}

\begin{figure}[!ht]
\centering
\includegraphics[scale=0.6]{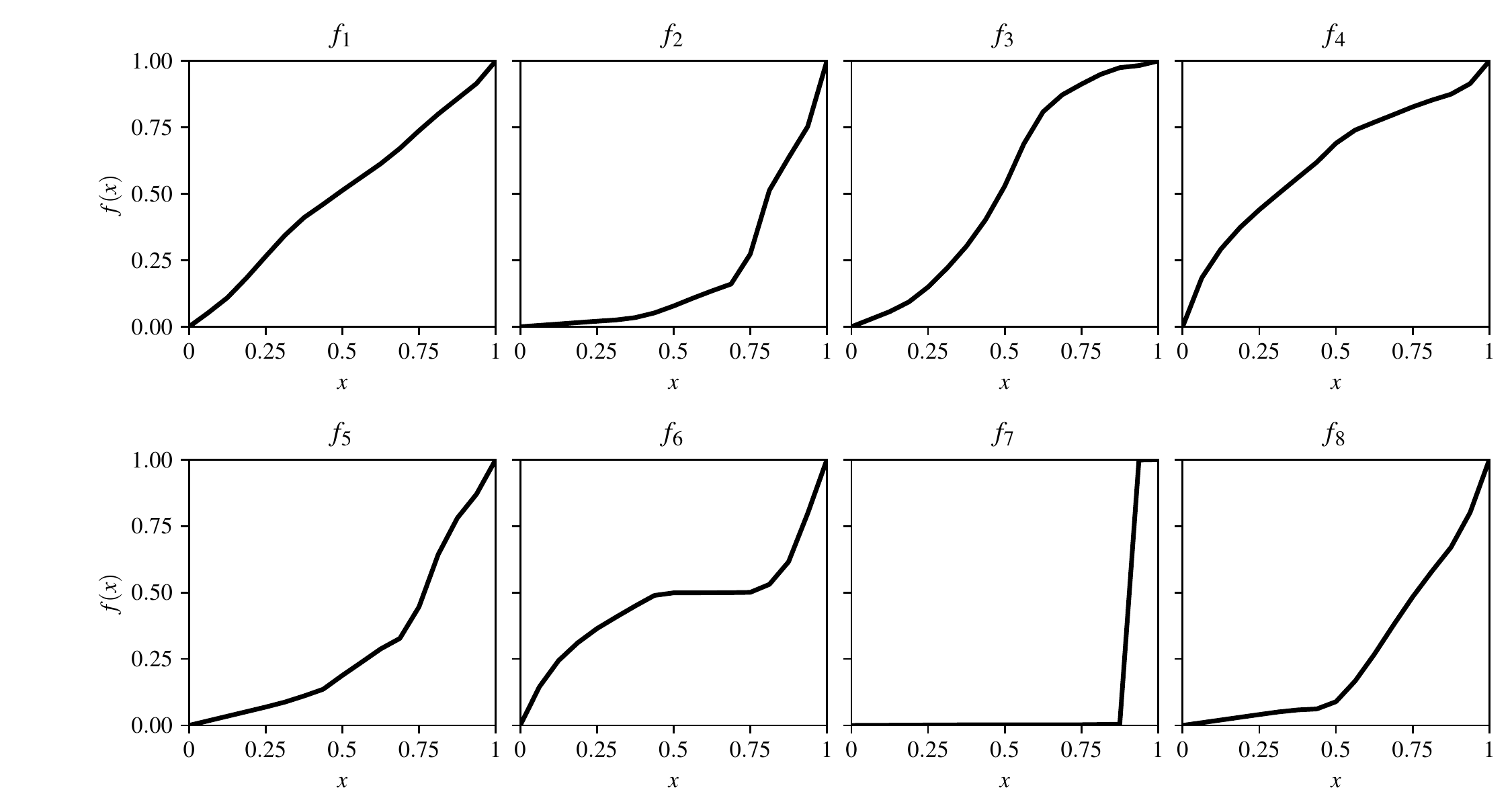}
\caption{Shape of activation functions for (Baseline + count) model with use of second glimpse}
\label{baseline_count_glimpse_f}
\end{figure}

\begin{figure*}[!ht]
    \centering
    \begin{subfigure}[b]{0.48\textwidth}
        \centering
        \includegraphics[width=\textwidth]{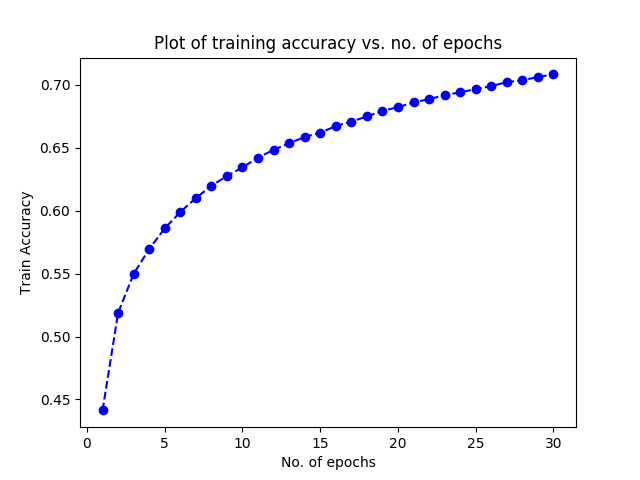}
        \caption[]%
        {{\small Plot of Training Accuracy for (Baseline + count) model}}    
    \end{subfigure}
    \hfill
    \begin{subfigure}[b]{0.48\textwidth}  
        \centering 
        \includegraphics[width=\textwidth]{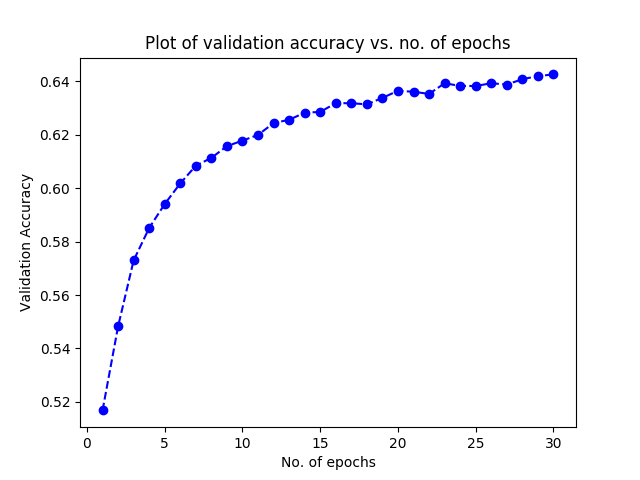}
        \caption[]%
        {{\small Plot of Validation Accuracy for (Baseline + count) model}}    
    \end{subfigure}
    \vskip\baselineskip
    \begin{subfigure}[b]{0.48\textwidth}   
        \centering 
        \includegraphics[width=\textwidth]{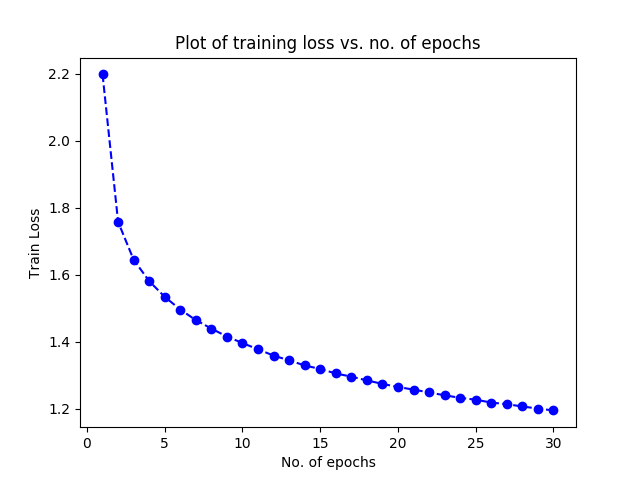}
        \caption[]%
        {{\small Plot of Training Loss for (Baseline + count) model}}    
    \end{subfigure}
    \quad
    \begin{subfigure}[b]{0.48\textwidth}   
        \centering 
        \includegraphics[width=\textwidth]{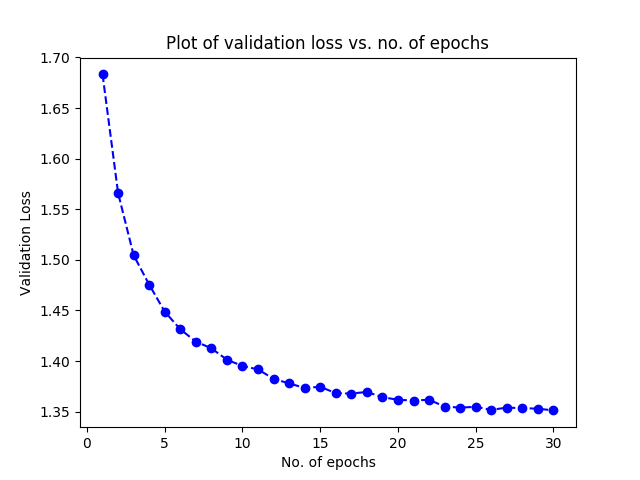}
        \caption[]%
        {{\small Plot of Validation Loss for (Baseline + count) model}}    
    \end{subfigure}
    \caption[]
    {\small Plot of Accuracy and Loss vs. no. of epochs for (Baseline+count model)} 
\end{figure*}
    

\begin{figure*}[!ht]
    \centering
    \begin{subfigure}[b]{0.48\textwidth}
        \centering
        \includegraphics[width=\textwidth]{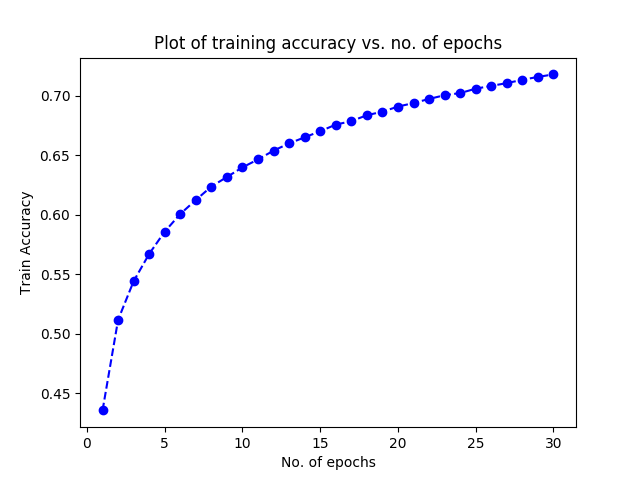}
        \caption[]%
        {{\small Plot of Training Accuracy for (Baseline count) model with unidirectional LSTM}}    
    \end{subfigure}
    \hfill
    \begin{subfigure}[b]{0.48\textwidth}  
        \centering 
        \includegraphics[width=\textwidth]{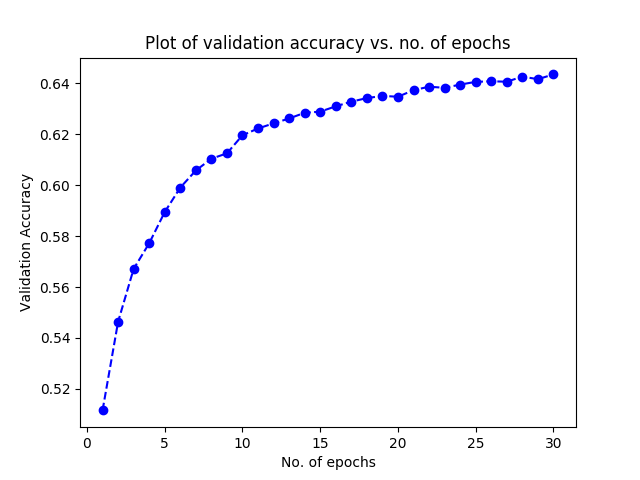}
        \caption[]%
        {{\small Plot of Validation Accuracy for (Baseline count) model with unidirectional LSTM}}    
    \end{subfigure}
    \vskip\baselineskip
    \begin{subfigure}[b]{0.48\textwidth}   
        \centering 
        \includegraphics[width=\textwidth]{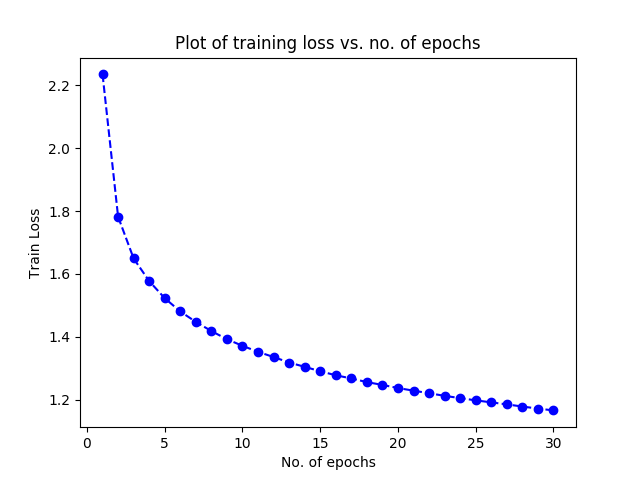}
        \caption[]%
        {{\small Plot of Training Loss for (Baseline count) model with unidirectional LSTM}}    
    \end{subfigure}
    \quad
    \begin{subfigure}[b]{0.48\textwidth}   
        \centering 
        \includegraphics[width=\textwidth]{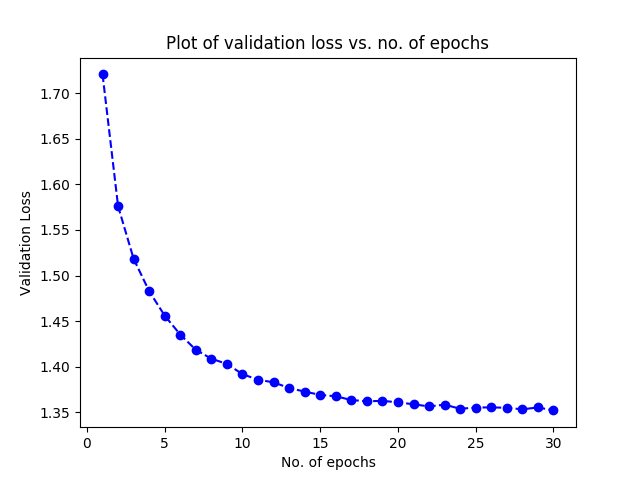}
        \caption[]%
        {{\small Plot of Validation Loss for (Baseline count) model with unidirectional LSTM}}    
    \end{subfigure}
    \caption[]
    {\small Plot of Accuracy and Loss vs. no. of epochs for (Baseline+count) model with unidirectional LSTM} 
\end{figure*}


\begin{figure*}[!ht]
    \centering
    \begin{subfigure}[b]{0.48\textwidth}
        \centering
        \includegraphics[width=\textwidth]{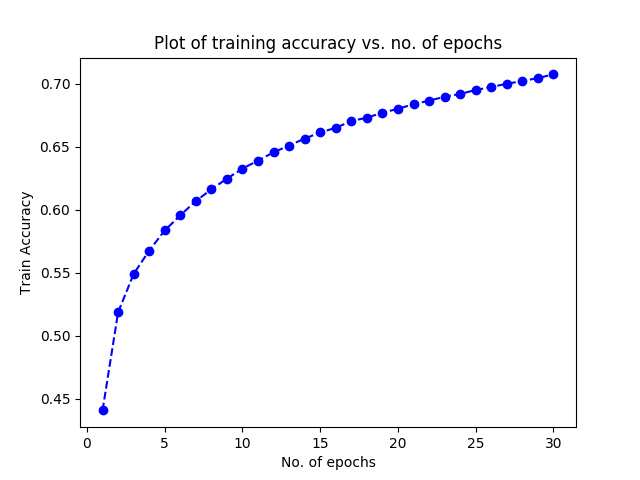}
        \caption[]%
        {{\small Plot of Training Accuracy for (Baseline count) model with $n=20$ objects}}    
    \end{subfigure}
    \hfill
    \begin{subfigure}[b]{0.48\textwidth}  
        \centering 
        \includegraphics[width=\textwidth]{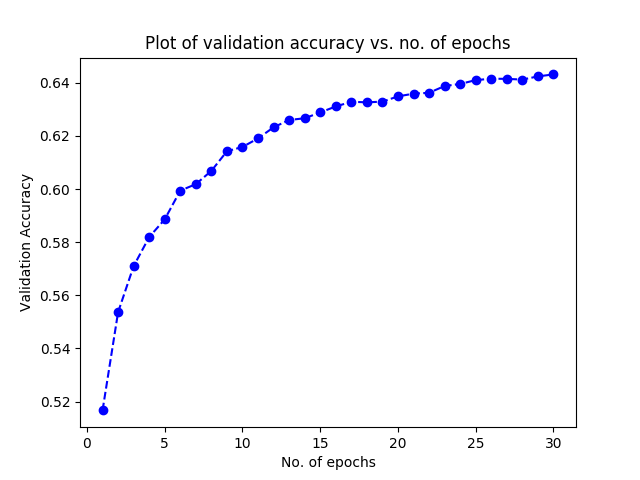}
        \caption[]%
        {{\small Plot of Validation Accuracy for (Baseline count) model with $n=20$ objects}}    
    \end{subfigure}
    \vskip\baselineskip
    \begin{subfigure}[b]{0.48\textwidth}   
        \centering 
        \includegraphics[width=\textwidth]{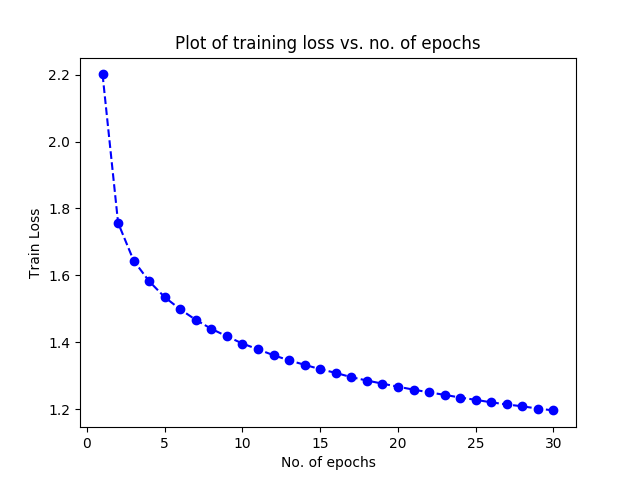}
        \caption[]%
        {{\small Plot of Training Loss for (Baseline count) model with $n=20$ objects}}    
    \end{subfigure}
    \quad
    \begin{subfigure}[b]{0.48\textwidth}   
        \centering 
        \includegraphics[width=\textwidth]{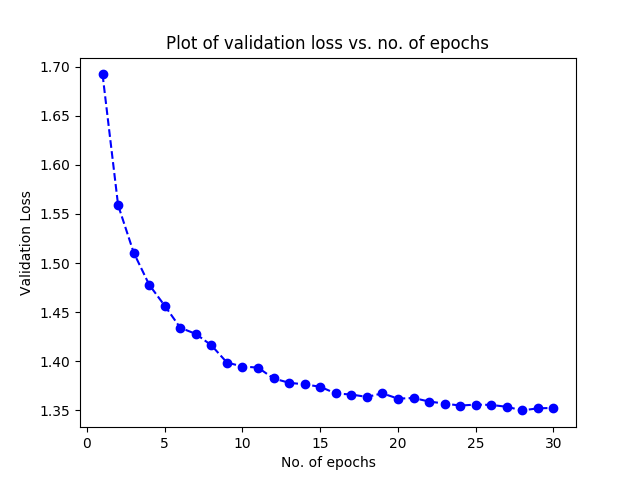}
        \caption[]%
        {{\small Plot of Validation Loss for (Baseline count) model with $n=20$ objects}}    
    \end{subfigure}
    \caption[]
    {\small Plot of Accuracy and Loss vs. no. of epochs for (Baseline+count) model with $n=20$ objects} 
\end{figure*}


\begin{figure*}[!ht]
    \centering
    \begin{subfigure}[b]{0.48\textwidth}
        \centering
        \includegraphics[width=\textwidth]{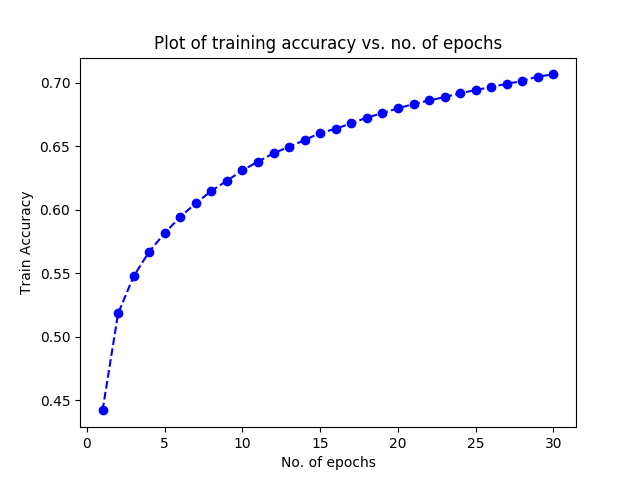}
        \caption[]%
        {{\small Plot of Training Accuracy for (Baseline count) model with confidence threshold 0.2}}    
    \end{subfigure}
    \hfill
    \begin{subfigure}[b]{0.48\textwidth}  
        \centering 
        \includegraphics[width=\textwidth]{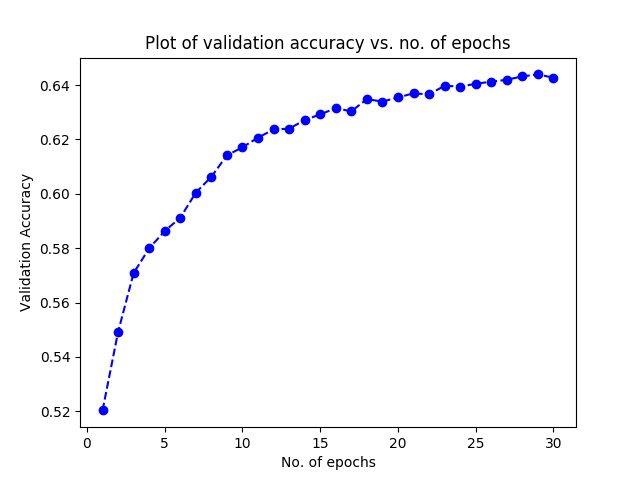}
        \caption[]%
        {{\small Plot of Validation Accuracy for (Baseline count) model with confidence threshold 0.2}}    
    \end{subfigure}
    \vskip\baselineskip
    \begin{subfigure}[b]{0.48\textwidth}   
        \centering 
        \includegraphics[width=\textwidth]{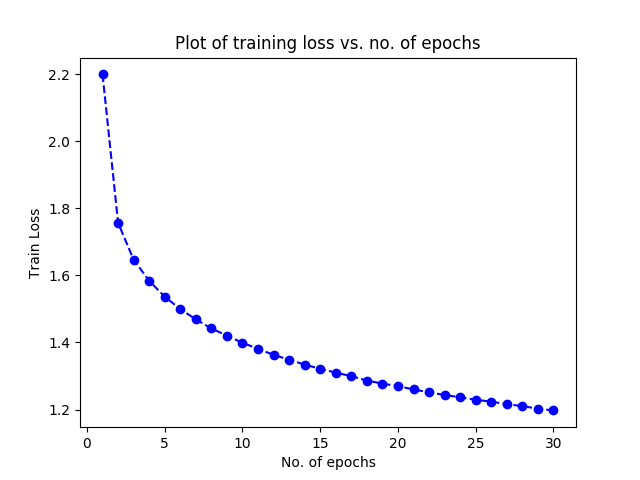}
        \caption[]%
        {{\small Plot of Training Loss for (Baseline count) model with confidence threshold 0.2}}    
    \end{subfigure}
    \quad
    \begin{subfigure}[b]{0.48\textwidth}   
        \centering 
        \includegraphics[width=\textwidth]{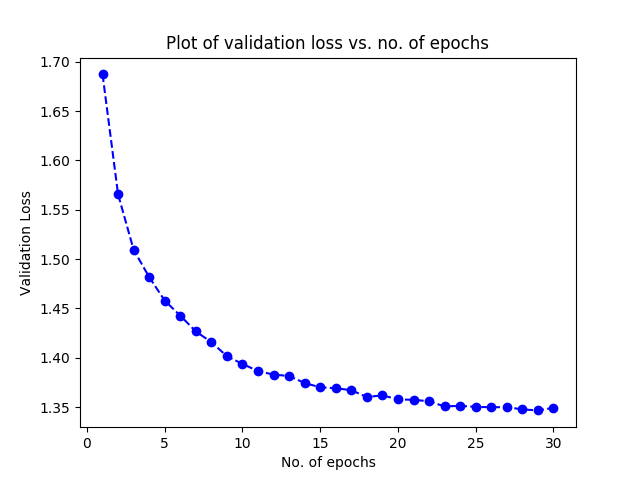}
        \caption[]%
        {{\small Plot of Validation Loss for (Baseline count) model with confidence threshold 0.2}}    
    \end{subfigure}
    \caption[]
    {\small Plot of Accuracy and Loss vs. no. of epochs for (Baseline+count) model with confidence threshold 0.2} 
\end{figure*}


\begin{figure*}[!ht]
    \centering
    \begin{subfigure}[b]{0.48\textwidth}
        \centering
        \includegraphics[width=\textwidth]{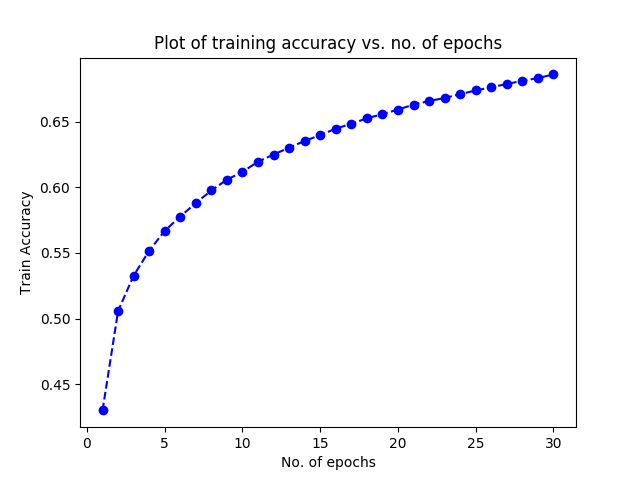}
        \caption[]%
        {{\small Plot of Training Accuracy for (Baseline count) model with embedding size 100}}    
    \end{subfigure}
    \hfill
    \begin{subfigure}[b]{0.48\textwidth}  
        \centering 
        \includegraphics[width=\textwidth]{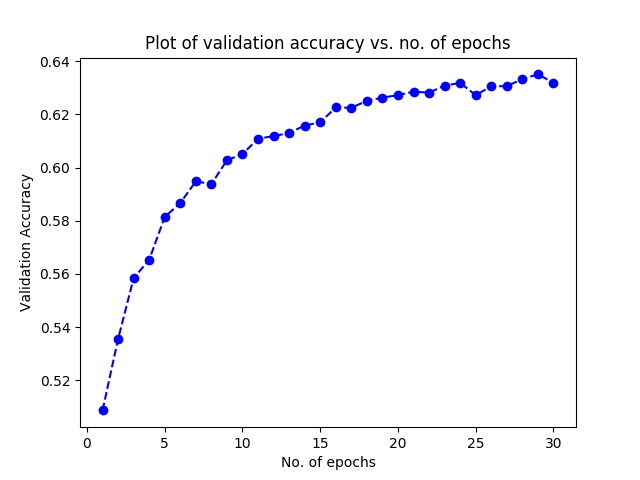}
        \caption[]%
        {{\small Plot of Validation Accuracy for (Baseline count) model with embedding size 100}}    
    \end{subfigure}
    \vskip\baselineskip
    \begin{subfigure}[b]{0.48\textwidth}   
        \centering 
        \includegraphics[width=\textwidth]{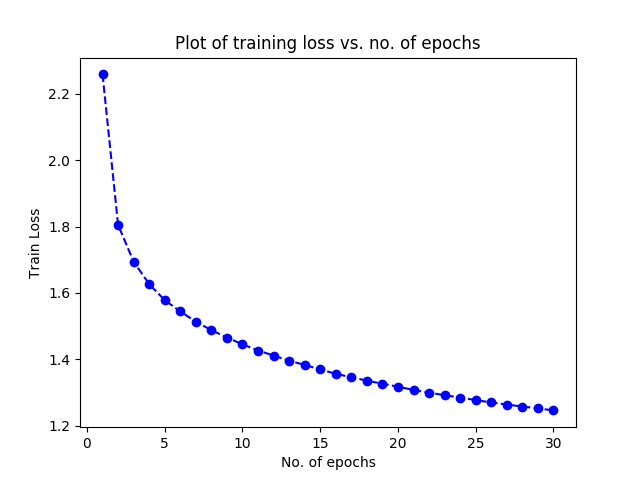}
        \caption[]%
        {{\small Plot of Training Loss for (Baseline count) model with embedding size 100}}    
    \end{subfigure}
    \quad
    \begin{subfigure}[b]{0.48\textwidth}   
        \centering 
        \includegraphics[width=\textwidth]{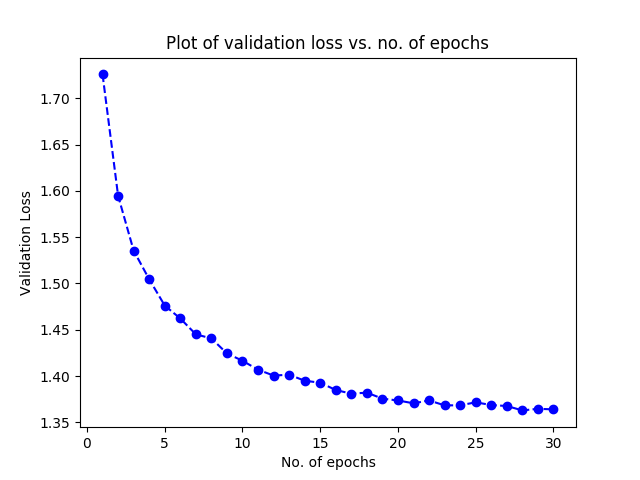}
        \caption[]%
        {{\small Plot of Validation Loss for (Baseline count) model with embedding size 100}}    
    \end{subfigure}
    \caption[]
    {\small Plot of Accuracy and Loss vs. no. of epochs for (Baseline+count) model with embedding size 100} 
\end{figure*}


\begin{figure*}[!ht]
    \centering
    \begin{subfigure}[b]{0.48\textwidth}
        \centering
        \includegraphics[width=\textwidth]{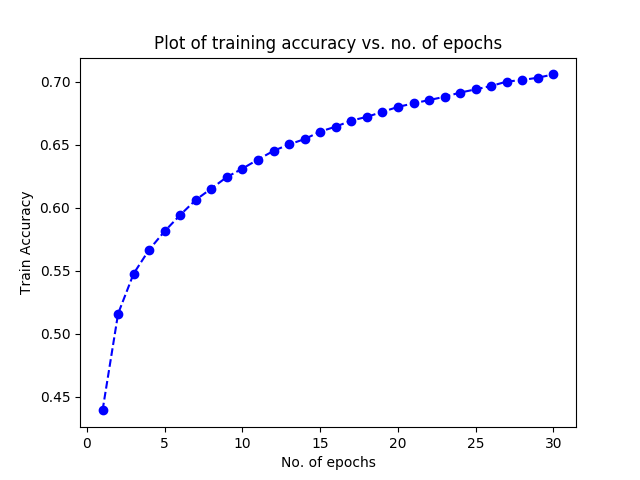}
        \caption[]%
        {{\small Plot of Training Accuracy for (Baseline count) model with use of second glimpse}}    
    \end{subfigure}
    \hfill
    \begin{subfigure}[b]{0.48\textwidth}  
        \centering 
        \includegraphics[width=\textwidth]{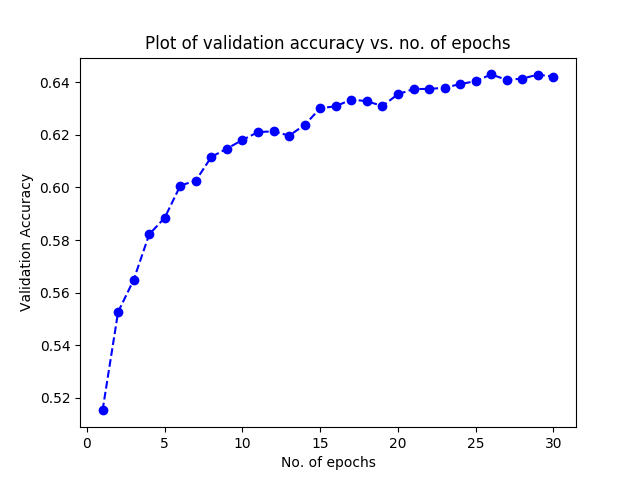}
        \caption[]%
        {{\small Plot of Validation Accuracy for (Baseline count) model with use of second glimpse}}    
    \end{subfigure}
    \vskip\baselineskip
    \begin{subfigure}[b]{0.48\textwidth}   
        \centering 
        \includegraphics[width=\textwidth]{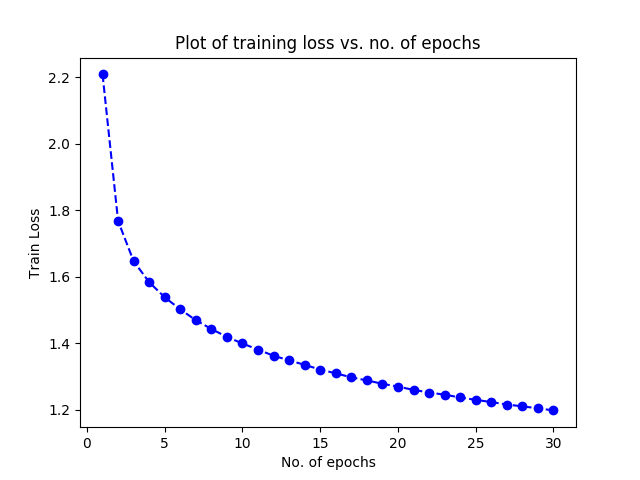}
        \caption[]%
        {{\small Plot of Training Loss for (Baseline count) model with use of second glimpse}}    
    \end{subfigure}
    \quad
    \begin{subfigure}[b]{0.48\textwidth}   
        \centering 
        \includegraphics[width=\textwidth]{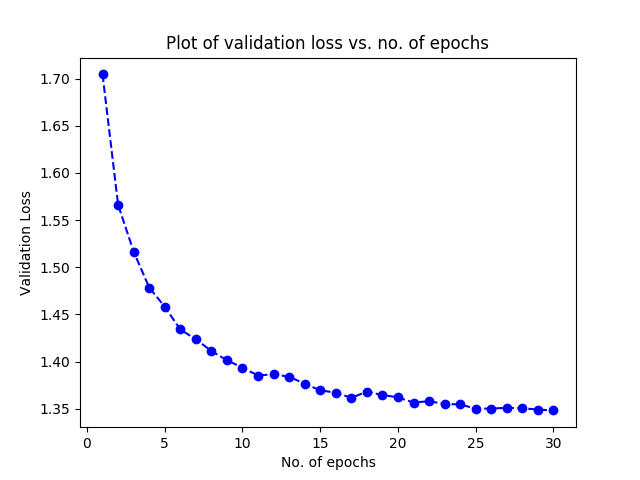}
        \caption[]%
        {{\small Plot of Validation Loss for (Baseline count) model with use of second glimpse}}    
    \end{subfigure}
    \caption[]
    {\small Plot of Accuracy and Loss vs. no. of epochs for (Baseline+count) model with use of second glimpse} 
\end{figure*}


\begin{figure*}[!ht]
    \centering
    \begin{subfigure}[b]{0.48\textwidth}
        \centering
        \includegraphics[width=\textwidth]{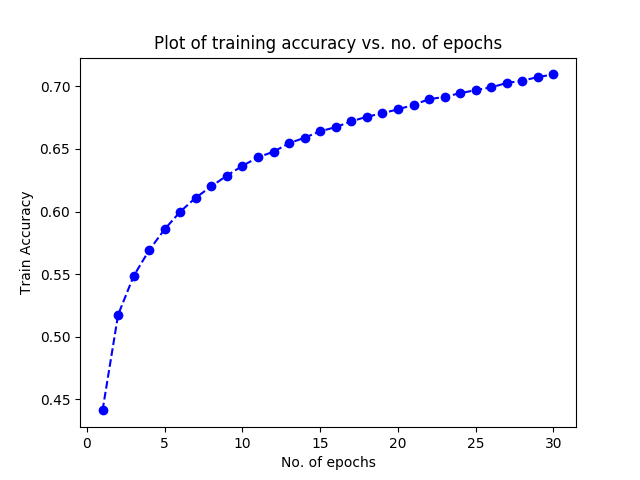}
        \caption[]%
        {{\small Plot of Training Accuracy for (Baseline) model}}    
    \end{subfigure}
    \hfill
    \begin{subfigure}[b]{0.48\textwidth}  
        \centering 
        \includegraphics[width=\textwidth]{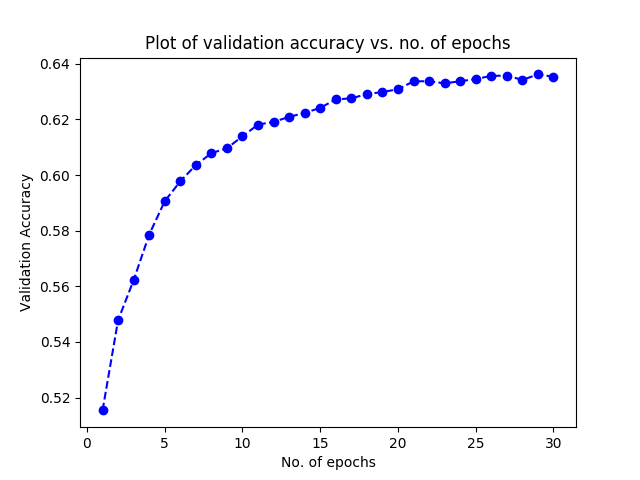}
        \caption[]%
        {{\small Plot of Validation Accuracy for (Baseline) model}}    
    \end{subfigure}
    \vskip\baselineskip
    \begin{subfigure}[b]{0.48\textwidth}   
        \centering 
        \includegraphics[width=\textwidth]{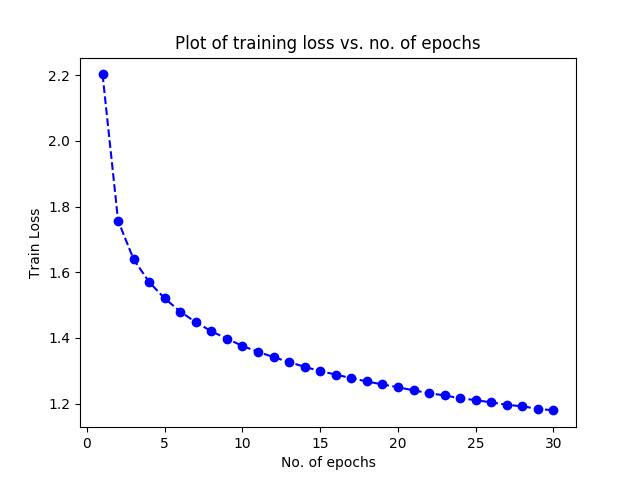}
        \caption[]%
        {{\small Plot of Training Loss for (Baseline) model}}    
    \end{subfigure}
    \quad
    \begin{subfigure}[b]{0.48\textwidth}   
        \centering 
        \includegraphics[width=\textwidth]{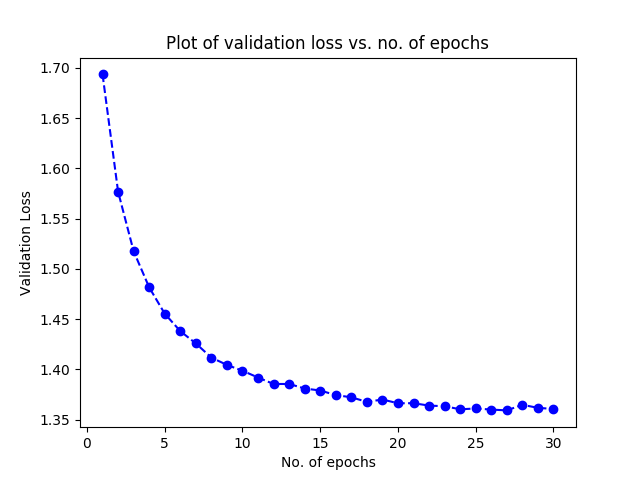}
        \caption[]%
        {{\small Plot of Validation Loss for (Baseline) model}}    
    \end{subfigure}
    \caption[]
    {\small Plot of Accuracy and Loss vs. no. of epochs for (Baseline) model} 
\end{figure*}


\begin{figure*}[!ht]
    \centering
    \begin{subfigure}[b]{0.48\textwidth}
        \centering
        \includegraphics[width=\textwidth]{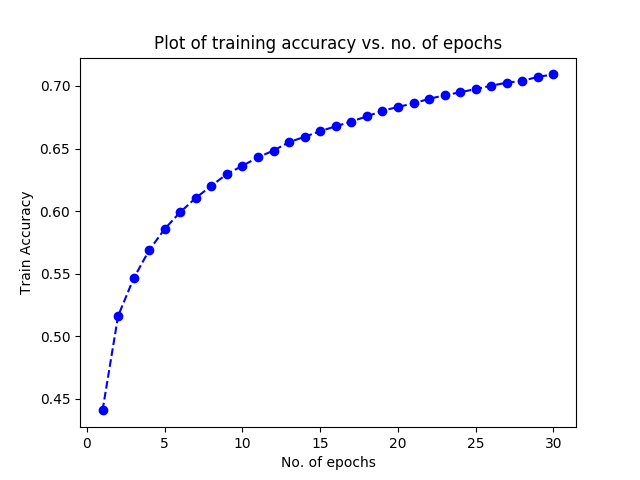}
        \caption[]%
        {{\small Plot of Training Accuracy for (Baseline) model with bidirectional LSTM}}    
    \end{subfigure}
    \hfill
    \begin{subfigure}[b]{0.48\textwidth}  
        \centering 
        \includegraphics[width=\textwidth]{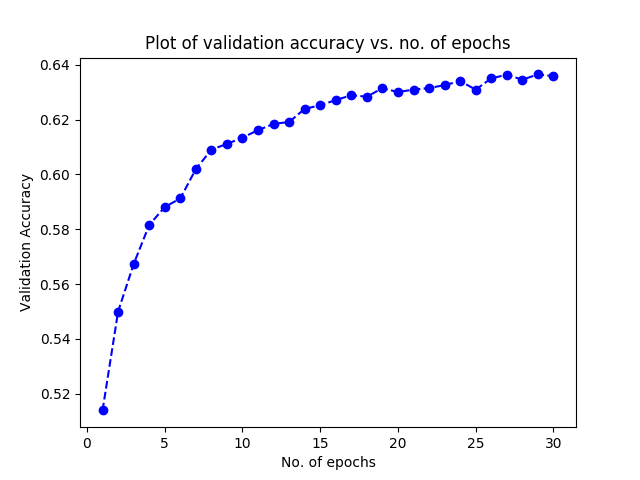}
        \caption[]%
        {{\small Plot of Validation Accuracy for (Baseline) model with bidirectional LSTM}}    
    \end{subfigure}
    \vskip\baselineskip
    \begin{subfigure}[b]{0.48\textwidth}   
        \centering 
        \includegraphics[width=\textwidth]{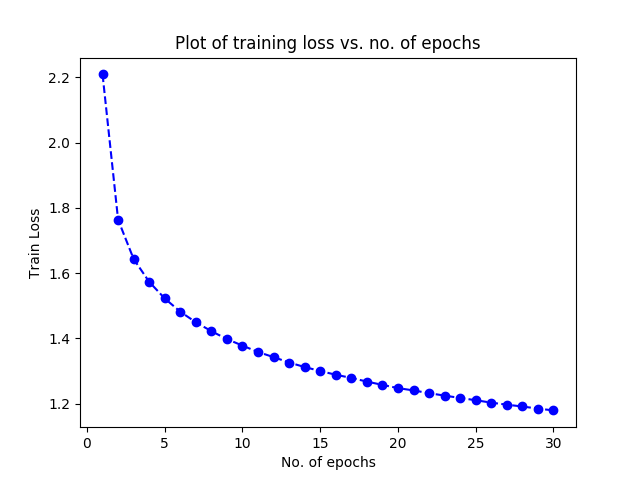}
        \caption[]%
        {{\small Plot of Training Loss for (Baseline) model with bidirectional LSTM}}    
    \end{subfigure}
    \quad
    \begin{subfigure}[b]{0.48\textwidth}   
        \centering 
        \includegraphics[width=\textwidth]{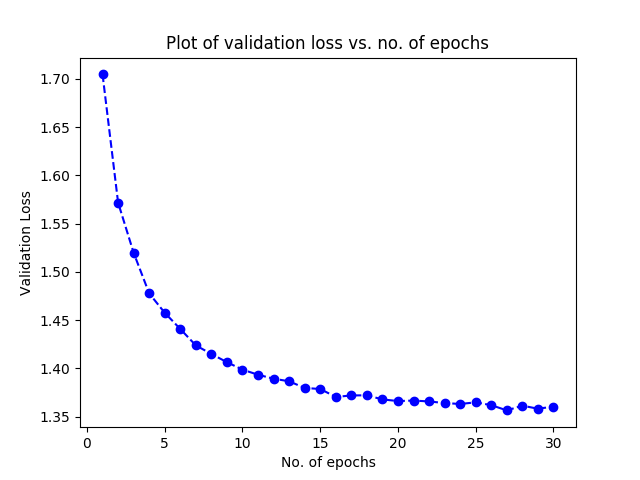}
        \caption[]%
        {{\small Plot of Validation Loss for (Baseline) model with bidirectional LSTM}}    
    \end{subfigure}
    \caption[]
    {\small Plot of Accuracy and Loss vs. no. of epochs for (Baseline) model} 
\end{figure*}

\end{document}